\documentclass[11pt,draftclsnofoot,onecolumn,journal]{IEEEtran}
\usepackage{amsmath}
\usepackage{amsfonts}
\usepackage{bm}
\usepackage{amsthm}
\usepackage{amssymb}
\usepackage{pgfplots}
\pgfplotsset{compat=1.18}
\usepackage{subcaption}
\usepackage{algorithm}
\usepackage{algpseudocode}
\usepackage{array}
\usepackage{textcomp}
\usepackage{stfloats}
\usepackage{url}
\usepackage{verbatim}
\usepackage{graphicx}
\usepackage{cite}
\usepackage{hyperref}
\usepackage{booktabs}
\usepackage{xcolor}   
\usepackage[utf8]{inputenc} 
\usepackage[T1]{fontenc}    
\usepackage{nicefrac}       
\usepackage{microtype}      
\usepackage{tcolorbox}
\usepackage{caption}
\usepackage{subcaption}
\usepackage{float}
\usepackage{tikz}
\usepackage{enumitem}
\usepackage{soul}
\usepackage{cancel}
\usepackage{bbm}
\usepackage{multirow}
\usepackage{makecell}
\usetikzlibrary{shapes, arrows.meta, positioning}

\newcounter{AFCounter}

\usepackage{cleveref}
\crefformat{equation}{(#2#1#3)}
\crefrangeformat{equation}{(#3#1#4) to~(#5#2#6)}
\crefmultiformat{equation}{(#2#1#3)}%
{ and~(#2#1#3)}{, (#2#1#3)}{ and~(#2#1#3)}

\newtheorem{theorem}{Theorem}

\newcommand{\abs}[1]{\left\vert #1 \right\vert}

\DeclareMathAlphabet{\mathsfit}{\encodingdefault}{\sfdefault}{m}{sl}
\SetMathAlphabet{\mathsfit}{bold}{\encodingdefault}{\sfdefault}{bx}{n}

\def\gE{{\mathcal{E}}}

\def\gH{{\mathcal{H}}}

\def\gL{{\mathcal{L}}}
\def\gM{{\mathcal{M}}}

\def\gS{{\mathcal{S}}}

\def\gX{{\mathcal{X}}}
\def\gY{{\mathcal{Y}}}

\newcommand{\E}{\mathbb{E}}
\newcommand{\N}{\mathbb{N}}

\newcommand{\R}{\mathbb{R}}

\newcommand{\set}[1]{\left\lbrace #1\right\rbrace}

\begin{document}
\title{Sustainable AI: Mathematical Foundations of Spiking Neural Networks}
\author{%
 \textbf{Adalbert Fono}$^{1}$, \textbf{Manjot Singh}$^{1}$, \textbf{Ernesto Araya},$^{1}$\\ \textbf{Philipp C. Petersen}$^{2}$, \textbf{Holger Boche}$^{3}$, \textbf{Gitta Kutyniok}$^{1,4,5,6}$    
 \vspace{0.25cm}\\
  $^{1}$ Department of Mathematics, Ludwig-Maximilians-Universität München, Germany \\
  $^{2}$ Faculty of Mathematics and Research Network Data Science, University of Vienna, Austria\\
 $^{3}$ Institute of Theoretical Information Technology, Technical University of Munich, Germany\\
  $^{4}$ Munich Center for Machine Learning (MCML), Munich, Germany\\
  $^{5}$ Department of Physics and Technology, University of Troms\o, Troms\o, Norway\\
  $^{6}$ Institute of Robotics and Mechatronics, DLR-German Aerospace Center, Germany
}

\maketitle

\begin{abstract}
    Deep learning's success comes with growing energy demands, raising concerns about the long-term sustainability of the field. Spiking neural networks, inspired by biological neurons, offer a promising alternative with potential computational and energy-efficiency gains. This article examines the computational properties of spiking networks through the lens of learning theory, focusing on expressivity, training, and generalization, as well as energy-efficient implementations while comparing them to artificial neural networks. By categorizing spiking models based on time representation and information encoding, we highlight their strengths, challenges, and potential as an alternative computational paradigm. 
\end{abstract}

\section{Introduction}
\label{secrel}

Deep learning has gained widespread attention and achieved state-of-the-art results in recent years across various domains, including image recognition, natural language processing, control systems, and logical reasoning \cite{applicatins2015_ylecun, Davies2021AdvancingMB}.
Recent breakthroughs have been driven by foundational models like GPT-4 \cite{openai2024gpt4technicalreport}, which achieved unprecedented results in their respective domains. This progress was driven by the expansion of training datasets and the increasing size and complexity of the underlying systems, the so-called \emph{artificial neural networks} (ANNs). Despite their unquestionable success, the enormous energy consumption of foundational models is a growing concern. For example, it is estimated that training GPT-4, with its approximately $1.75$ trillion parameters, 
consume 
megawatt-hours of electricity---equivalent to 
the annual consumption of $1300$ average American households.
This need for vast resources raises questions about the long-term sustainability of the current trajectory of ANN development \cite{Thompson2021DimReturns}. 
Alongside the high energy demands, ANNs exhibit other drawbacks, such as stability issues, a gap between theory and practice in their generalization abilities, and challenges with explainability, to name a few. The persistence of these deficiencies, even in the latest models, suggests that they may be intrinsic to the current ANN paradigm. As a result, ongoing improvements alone may not sufficiently address these challenges, indicating that more efficient computational methods might be required.

The implementation of \emph{spiking neural networks} (SNNs) on neuromorphic hardware provides a promising alternative to the current computing paradigm. The term ‘neuromorphic’ originally described systems emulating specific aspects of biological neural systems. Now, it more broadly refers to systems manifesting brain-inspired properties including fine-grained parallelism, reduced precision computing, and in-memory computing. These properties are already exploited in contemporary neuromorphic chips but are expected to yield further gains with continuing development \cite{Mehonic2024NeuromorphicRoadmap}. 
The key innovation of SNNs adopted from biological systems---the communication between neurons through asynchronous discrete electrical pulses or \emph{spikes}---marks a clear distinction to traditional ANNs with synchronous information propagation \cite{gerstner_kistler_naud_paninski_2014}. After first proposals in the 80s to employ spike-based systems as a computing model, e.g., by Hopfield \cite{Hopfield82NNs}, Maass provided in the 90s a rigorous mathematical analysis of the computational power of spiking neurons. 
He also coined SNNs as the `third generation' of neural networks, 
succeeding 
the perceptron and sigmoidal models \cite{Maass1996Networksthirdgeneration}. The goal of developing a more biologically realistic neural model is to mimic the remarkable computational and energy efficiency of the human brain. Despite containing approximately 100 billion neurons and $10^{14}$ synapses—100 times more `parameters' than GPT-4—the brain operates on just 20 watts of power.
This led to a resurgence of SNNs in recent years with promising results, however, a clear understanding of their full capacities is still lacking.

Developing a comprehensive mathematical theory would deepen our understanding of SNNs and 
facilitate the innovations from 
diverse research directions.
However, the theoretical properties of SNNs remain under-explored in the broader AI community. 
Research on SNNs has primarily focused on developing learning algorithms to achieve competitive performance in real-world applications. This emphasis is due to the unique challenges presented by spike dynamics, where the learning objective is a non-differentiable function, rendering typical gradient-based tools insufficient \cite{Guo_direct_training_review_2023, Rathi_neuromorphic_based_SNNs_2023, SNNSurvey2023}. 
We aim to broaden the scope by presenting a thorough and accessible overview of the theoretical landscape of SNNs. 
A main theme throughout the survey is to draw parallels with the established mathematical theory of ANNs and highlight topics central to SNNs not covered by the classical theory, in particular, questions regarding energy consumption. To that end, we will first introduce the computational framework describing SNNs in \Cref{secpr}, followed by its analysis regarding expressivity, training, generalization, and energy-efficiency in the subsequent Sections \ref{subsection:expressivity} - \ref{sec:energy}.

\section{Foundations of Spiking Neural Networks}
\label{secpr}
The landscape describing the dynamics of (biological) spiking neurons is quite diverse and encompasses various mathematical models, ranging from complex (Hodgkin-Huxley) to rather simplified (Integrate-and-Fire) \cite{gerstner_kistler_naud_paninski_2014}. 
The envisioned use case might not only favor a particular model but also influence certain design choices within the model. 
Broadly speaking, SNNs can be applied in neuroscience to 
improve our understanding of 
the brain, or SNNs can be considered as a computational model progressing towards robust and efficient AI \cite{mineault2024neuroaiaisafety}. This survey focuses on the latter approach that intrinsically leans towards simpler mathematical models, 
which can be efficiently implemented and employed on computing platforms.
The main innovation introduced by SNNs as computational models is their spike-based processing. Therefore, the key question is whether the leap from classical computational neurons to spiking neurons is sufficient to realize the benefits promised by biological neural networks or more biological features need to be captured by the models first.
At a high level, replacing classical artificial neurons with spiking neurons in a network results in the following characteristics:
(i) information processing in SNNs is event-based/asynchronous, in contrast to the sequential/synchronous processing in ANNs, (ii) spikes are discrete events in time, meaning they propagate binary information (spike or no spike at a given moment) as long as the shape, duration, etc., of the spike do not carry additional information (which is typically the case in computational models), in contrast to the analog information—i.e., real values—used in ANNs, 
and (iii) spiking neurons have a non-constant state (spiking vs. non-spiking) in contrast to classical neurons.
What impact do these properties have on the performance and potential of both ANNs and SNNs? Due to the binary information propagation, SNNs appear to be more limited than ANNs, however, the event-based nature of spikes and the statefulness of neurons could be seen as enhancing the computational capacity of SNNs in certain circumstances. Another layer of complexity is added by mechanisms for encoding information in spikes, e.g., via their frequency or timing.  
Therefore, we must assess how these properties are implemented in computational models to obtain more qualitative statements. Since most models allow for many optional design choices it is not straightforward to analyze the impact of the described properties on their computational power/efficiency.  
Hence, we will focus on two key aspects, namely the handling of the time dimension (discrete or continuous) as well as the type of information propagation (discrete or continuous), and classify models along these two axes to study the similarities and differences of the classes; we will return to this aspect in \Cref{sec:Motivation} after introducing the neuronal dynamics.

\subsection{Models of neuronal dynamics}
In this survey, we study two of the most prominent and heavily employed computational models of spiking neurons, namely Integrate-and-Fire models and the Spike Response Model. This choice is motivated by their proven effectiveness and simplicity, allowing us to focus on the core aspects of this computational paradigm.
The same reasoning also guides us toward simple network structures and a deterministic setting, while acknowledging that more elaborated structures, as well as stochastic frameworks, could enhance their computational capacity \cite{Noisyspikingneurons_Maass1996}. In the remainder, 
we focus on mathematical frameworks, neglecting biological interpretation; for a detailed biological background on spiking neurons, see \cite{gerstner_kistler_naud_paninski_2014}.

\subsubsection{Models of biological neurons} One of the most basic models of neuronal dynamics are \emph{Integrate-and-Fire} (IF) models that generally consist of two components: (i) The description of the evolution of the \emph{(membrane) potential} of a neuron and (ii) the spike generation mechanism. 
The simplest IF model is the \emph{Leaky-Integrate-and-Fire} (LIF) model, where state evolution follows a linear differential equation, and spike generation occurs via a thresholding operation.
The governing differential equation is inspired by a resistor-capacitor circuit driven by a current $I$ evolving over time
\begin{equation}\label{eq:LIF1}
   \tau_m \frac{\mathrm{d} u}{\mathrm{d} t}(t) = - (u(t) - u_\text{rest}) + I(t) \qquad \text{ for } t > t_0,       
\end{equation}
where $u$ denotes the \emph{(membrane) potential}, $u_\text{rest}$ the \emph{resting potential}, and $\tau_m >0$ the \emph{membrane time constant}. 
The solution of \eqref{eq:LIF1} with the initial condition $u(t_0) = u_\text{rest} + \Delta u$, $\Delta u \in \R$, is given by
\begin{equation}\label{eq:TDSol}
    u(t) = u_\text{rest} + \frac{1}{\tau_m} \int_0^{t-t_0} \exp{(-\frac{s}{\tau_m})} I(t-s) \mathrm{d} s + \Delta u \exp{(-\frac{t-t_0}{\tau_m})} \qquad \text{ for } t > t_0.  
\end{equation}
Assuming that the neuron is in the resting state, i.e., $u_\text{rest}$ coincides with $u(t_0)$, (or equivalently assuming that $t_0 \to - \infty$) gives the form
\begin{equation}\label{eq:TDSol1}
    u(t) = u_\text{rest} + \frac{1}{\tau_m} \int_0^\infty \exp{(-\frac{s}{\tau_m})} I(t-s)\mathrm{d} s \qquad \text{ for } t \in \R.
\end{equation}
Spikes in the LIF model are introduced by fixing a threshold $\vartheta >0$ on the potential, i.e., the neuron emits a spike at time $t^f$ if $u(t^f)=\vartheta$. The actual shape of the spike (expressed in terms of the potential) is neglected but the firing time is noted and immediately after the spike the potential is reset to a new value $u_r < \vartheta$ such that 
$\lim_{\varepsilon \searrow 0} u(t^f + \varepsilon) = u_r$.
Spikes are thus reduced to points in time and we denote the sequence of firing times of a neuron, the so-called \emph{spike train}, as $S(t)=\sum_i \delta(t-t^{f_i})$,
where $\delta$ denotes the Dirac delta distribution and $f_i$ for $i=1,2,\dots$ is the label of the spike. 

Returning to our electrical circuit analogy, the reset of the potential corresponds to a short current pulse $I_r(t)= - \tau_m (\vartheta - u_r)\delta(t - t^f)$ at firing time $t^f$. In the general case of a spike train $I_r(t)=- \tau_m(\vartheta - u_r) S(t)$, inserting the total current $I(t) + I_r(t)$ in \eqref{eq:LIF1} yields the complete differential description 
\begin{equation}\label{eq:FullLIFDE}
     \frac{\mathrm{d} u}{\mathrm{d} t}(t) = - \frac{1}{\tau_m}\big((u(t) - u_\text{rest}) - (I(t) +I_r(t)\big) =  - \frac{1}{\tau_m}\big((u(t) - u_\text{rest}) - I(t)\big) + (u_r - \vartheta) S(t) \text{ for } t > t_0 
\end{equation}
of the evolution of the potential of a LIF neuron with the corresponding solution derived via \eqref{eq:TDSol1} given as
\begin{align}\label{eq:GenSol}
    u(t) &= u_\text{rest} + (u_r - \vartheta)\sum_i \exp{(-\frac{t - t^{f_i}}{\tau_m})} +  \int_0^\infty \overbrace{\frac{1}{\tau_m}\exp{(-\frac{s}{\tau_m})}}^{=\kappa(s)} I(t-s) \mathrm{d} s \\
    &= u_\text{rest} + \int_0^\infty \eta(s) S(t-s) \mathrm{d} s +  \int_0^\infty \kappa(s) I(t-s) \mathrm{d} s, \quad\!\! \text{ for } t \in \R \text{ with } \eta(s) = (u_r - \vartheta) \exp{(-\frac{s}{\tau_m})} . \notag
\end{align}
We interpret $\eta$ and $\kappa$ as describing the potential reset and summarizing its linear electrical properties, respectively. Ignoring exponential specifics, we assume $\eta$ to represent the effect of an outgoing spike on the potential, while $\kappa$ captures its linear response to an incoming spike. 
This perspective leads to an alternative description of neuronal dynamics, the \emph{Spike Response Model} (SRM), based on the response kernels $\eta$ and $\kappa$. 
A firing occurs at time $t^f$ in the SRM if $u(t^f)= \vartheta(t^f)$ and $\frac{\mathrm{d}(u(t^f) - \vartheta(t^f)}{\mathrm{d} t} >0$, i.e., the threshold $\vartheta(t)$ is, in contrast to the LIF model, not fixed but time-dependent. 
The condition that the potential $u$ reaches the dynamic threshold from below is necessary because, unlike the LIF model's instantaneous reset, the reset mechanism follows a trajectory encoded in $\eta$.
The dynamic threshold in the SRM is meaningful as it allows the incorporation of neuronal refractoriness after spikes, in various ways.
As a side effect, $\eta$ can be integrated into the dynamic threshold by appropriately increasing the threshold at firing time. With $\eta$ incorporated in $\vartheta$, \eqref{eq:GenSol} simplifies to
\begin{equation}\label{eq:SRMSimple}
    u(t)= u_\text{rest} + \int_0^\infty \kappa(s) I(t-s) \mathrm{d} s \qquad \text{ for } t \in \R.  
\end{equation}

\subsubsection{Spiking neurons as computational models} 
Having established two models describing spiking neurons it is left to turn them into viable computational frameworks, i.e., biological plausibility is now secondary to computational efficiency. To employ networks of spiking neurons, we assume that the input current $I_i(t)$ of a neuron $i$ in the network at time $t$ is generated by presynaptic neurons via $I_i(t)= \sum_j \sum_k w_{i,j} \alpha(t-t^{f_k}_j)$, where $w_{i,j}$ and $\alpha(t-t^{f_k}_j)$ describe the weight factor of the synapse from $j$ to $i$ and the time course of the incoming synaptic current pulse from neuron $j$, respectively. Crucially, the weights $w_{i,j}$ may act as learnable parameters in a computational setting just as in the classical ANN framework. 
For this specific input current, the following expression, derived from \eqref{eq:SRMSimple}, describes the potential $u_i$ of an SRM neuron 
\begin{equation}\label{eq:SRMAlt}
    u_i(t) = u_\text{rest} + \sum_j \sum_k w_{i,j} \underbrace{\int_0^\infty \kappa(s)   \alpha(t-s-t^{f_k}_j) \mathrm{d} s}_{=: \varepsilon_{i,j}(t-t^{f_k}_j)}
    = u_\text{rest} + \sum_j \sum_k w_{i,j}  \varepsilon_{i,j}(t-t^{f_k}_j) \quad \text{ for } t\in\R,     
\end{equation}
which can be iterated to display the potential and spike time(s) of each neuron by solving
\begin{equation}\label{eq:SRMST}
    t_i^{f_k} = \min\{ t\in \R: t > t_i^{f_{k-1}} \text{ and } u_i(t) = \vartheta(t)\}.
\end{equation}
Instead of relying on a fixed response function $\varepsilon_{i,j}$, it is commonly considered an adjustable parameter to allow for more flexibility in this framework. Thereby, the choices range from biologically plausible and complex to less realistic but highly simplified opening up avenues for practical and theoretical treatment. In the latter case, an additional learnable parameter is often introduced for each synapse, the \emph{synaptic delay} $d_{i,j} \geq 0$ representing the transmission time of spikes via the synapse $j$ to $i$ (which is not directly accounted for in the response due to the simplification), by substituting $t^{f_k}_j$ with $t^{f_k}_j + d_{i,j}$ in \eqref{eq:SRMAlt}. 

A similar framework can be derived for LIF neurons by analogous consideration. 
Informally speaking, the framework enables us, in principle, to iteratively compute or approximate the firing times and to retrace the resulting spike pattern in the network.
The key feature is that the time parameter $t$ is treated as a continuous variable. However, for digital implementation purposes, this has certain downsides/overhead related to optimizing the learnable parameters due to the asynchronous propagation of spikes (see \Cref{section:training}).
To avoid these issues and align closer with the training pipeline of synchronous ANNs, where computations are performed sequentially layer-wise, the time dimension can be discretized which we demonstrate for LIF neurons yielding the \emph{discretized LIF} model. 
We neglect the reset mechanism first and convert the differential equation in \eqref{eq:LIF1} into a difference equation via the forward Euler method with time steps $t_n$, $n\in\N$, and step size $\Delta t >0$ (where we assumed w.l.o.g. that $u_\text{rest}=0$ and $t_0=0$ for clarity)
\begin{equation}\label{eq:Euler1}
    \tau_m \frac{u(t_{n+1}) - u(t_{n})}{\Delta t} = - u(t_n)  +  I(t_n) \iff u(t_{n+1})  = (1 - \frac{\Delta t}{\tau_m}) u(t_n)  + \frac{\Delta t}{\tau_m} I(t_n).
\end{equation}
Let $\beta >0$ denote the decay rate of the potential, i.e., the ratio between subsequent values of $u$ separated by $\Delta t$ in the absence of input currents, which can be explicitly computed via \eqref{eq:TDSol} with $I(t)=0$:
\begin{equation*}
    \beta = \frac{u(t_{n+1})}{ u(t_{n})} = \frac{u(t_0)  \exp{(-\frac{t_{n+1}-t_0}{\tau_m})} }{u(t_0) \exp{(-\frac{t_{n}-t_0}{\tau_m})}} = \exp{(-\frac{\Delta t}{\tau_m})}. 
\end{equation*}
Hence, we observe that the discretization process introduced a first-order approximation $1 - \tfrac{\Delta t}{\tau_m}$ of $\beta$ (via its series expansion) in \eqref{eq:Euler1}.  
Substituting the exact value of $\beta$ and time-shifting the input current by one step to allow for its instantaneous contribution to the potential yields the refined discretization
\begin{equation*}
    u(t_{n+1})  = \beta u(t_n)  + (1-\beta) I(t_{n+1}).        
\end{equation*}
To incorporate the reset mechanism, recall from the continuous case that the potential is reset after a spike and it returns over time to  $u_{\text{rest}}$ if no other spike or input current is registered (see \eqref{eq:FullLIFDE} and \eqref{eq:GenSol}). A natural way to convert this into the discretized setting is to reset the potential after the occurrence of a spike, indicated by $s(t_{n+1}) \in \{0,1\}$, directly to $u_{\text{rest}}$ and neglect the relaxation aspect of the potential. 
A common alternative is to employ the reset-by-subtraction method where instead of resetting the potential to zero only a certain amount, typically the threshold $\vartheta$, is subtracted from the potential:
\begin{equation}\label{eq:LIFDis}
    u(t_{n+1}) =  \beta u(t_n) + (1-\beta) I(t_{n+1}) - s(t_{n+1}) \vartheta.
\end{equation}
The two approaches converge for small step sizes, however, reset-by-subtraction is considered superior for implementation purposes as it retains residual superthreshold information by design \cite{SNNSurvey2023}.
As in the SRM, assume that the input current $I_i(t_{n}) =\sum_j w_{i,j} s_j(t_n)$ of a neuron $i$ is generated by spikes of presynaptic neurons so that \eqref{eq:LIFDis} can be rewritten via the Heaviside function $H$ into the final version of the model
\begin{equation}\label{eq:LIFDisFinal}
    \begin{cases}
        s_i(t_{n+1}) &= H(\beta u_i(t_n) + \sum_j w_{i,j} s_j(t_{n+1}) - \vartheta) \\
        u_i(t_{n+1}) &=  \beta u_i(t_n) + \sum_j w_{i,j} s_j(t_{n+1}) - s_i(t_{n+1})\vartheta    
    \end{cases},           
\end{equation}
since by construction a spike is emitted at time $t^f$ if $u(t^f) = \vartheta$ is satisfied and the coefficient $(1-\beta)$ is subsumed into the learnable weights $w_{i,j}$. To summarize, the benefit of discretization is that the derived model neatly fits into the training framework of ANNs despite some obstacles discussed in \Cref{section:training}.

\subsection{Networks of spiking neurons}
To explicitly model networks of neurons, it is typically assumed that the neurons form a graph structure, i.e., neurons and synapses represent vertices and (weighted) edges, respectively. In feedforward networks, which are the basis of more advanced structures such as convolutional, recurrent, etc., the underlying directed, acyclic graph is arranged in layers. Due to their fundamental importance and their elementary structure, they are the mathematically best understood ANN. Therefore, the feedforward setting is also a reasonable starting point for analyzing SNNs broadly characterized by the following properties: 
\begin{enumerate}[label=(\roman*)]
    \item \emph{Network spatial architecture} $(L,n) \in \N\times\N^{L+1}$, where $L$ is the number of hidden layers and $n=(n_0, \hdots, n_{L})$ is the number of neurons in each layer.     
    \item \emph{(Hyper)parameter} $\big( W^\ell, P^\ell \big)_{\ell\in [L]}$, where the \emph{weight matrices} $W^{\ell} \in \R^{n_\ell\times n_{\ell-1}}$ correspond to the weighted edges of the network graph and represent the learnable parameters also common in ANNs, whereas $P^\ell$ denote the (hyper)parameter specific to an SNN framework. 
    \item \emph{Neuronal (temporal) dynamics.} For each layer $\ell\in  [L]$, the evolution of the potential of its neurons depends on the incoming spikes from the previous layer and follows the dynamic described by the chosen model with weights $W^{\ell}$ and other parameter specified via $P^\ell$.
\end{enumerate}

The model-specific parameter $\varepsilon$ (response function) and $\vartheta$ (threshold function) of continuous-time SRM neurons \eqref{eq:SRMAlt} are typically shared throughout the network and, together with the (optional) synaptic delay matrices $D^\ell \in \R_{\geq 0}^{n_\ell \times n_{\ell-1}}$ representing synaptic transmission delays, reflect the propagation of spikes given initial spike times in the input layer according to \eqref{eq:SRMST}. 
In contrast, in the discretized LIF model in \eqref{eq:LIFDisFinal} the initial potential $u(0)\in \R$, the leaky term $\beta \in [0,1]$, and the threshold $\vartheta \in (0,\infty)$ of each neuron as well as the number of time steps $T\in \N$, which is assumed to be constant throughout the network, comprise the model specific parameter. Given an initial spike pattern $(s^0(t))_{t\in[T]} \in \set{0,1}^{n_0\times T}$ in the input layer, the layer-wise dynamics of the discretized LIF model according to \eqref{eq:LIFDisFinal} are 
\begin{equation}\label{eq:LIFSNN}
    \begin{cases}
        s^{\ell}(t) &= H\big(\beta^\ell u^{\ell}(t-1) + W^{\ell} s^{\ell-1}(t) + b^\ell- \vartheta^{\ell} \bm{1}_{n_{\ell}} \big)\\
        u^{\ell}(t) &= \beta^\ell u^{\ell}(t-1) + W^{\ell} s^{\ell-1}(t) + b^\ell- \vartheta^{\ell} s^{\ell}(t)  
    \end{cases},   
    \qquad \text{ for } t \in [T], \ell \in [L], 
\end{equation} 
where $H$ is applied entry-wise. In both cases an SNN $\Phi = \big( W^\ell, P^\ell \big)_{\ell\in [L]}$, characterized by its parameters and the underlying dynamics, \emph{realizes} the input to output mapping $R(\Phi): (\R^\N)^{n_0} \to (\R^\N)^{n_L}$.

\subsubsection{From Spike Patterns to Information Processing }
The final step in developing a complete computational model of SNNs is defining input and output representations. Although the introduced framework is quite flexible, the realizations operate in the spike domain, whereas tasks like classification typically require outputs outside the spike domain. This requires a bridge between the spike and task domains, similar to how biological neural networks convert spike-based information into actions. Various conversion mechanisms, which we will refer to as \emph{coding schemes} in the remainder, exist depending on whether an instantaneous or time-delayed reaction is required \cite{gerstner_kistler_naud_paninski_2014, MAASS2001ontherelevanceoftime}. 
For our purposes, the coding scheme can be thought of as two additional layers of an SNN $\Phi$ defined by the \emph{input encoding} $E: \R^{n_\text{in}} \to (\R^\N)^{n_0}$ and \emph{output decoding} $D: (\R^\N)^{n_L} \to \R^{n_{\text{out}}}$, where $n_\text{in}, n_{\text{out}}$ denote the input and output dimension of the given problem, respectively, i.e., $\Psi = (\Phi, (E,D))$ realizes the mapping $R(\Psi): \R^{n_\text{in}}\to \R^{n_{\text{out}}}$ given by $R(\Psi) = D \circ R(\Phi) \circ E$. 
From a computational perspective, an ideal coding scheme should strike the balance between practical implementation, theoretical soundness, and task performance. 
Practically, it should be easy to implement, robust, and facilitate efficient learning by either being differentiable or compatible with surrogate gradient methods, enabling seamless integration into typical learning pipelines. In the remainder, we focus on two predominant coding paradigms observed in both computational and biological systems---temporal and rate coding---highlighting their distinct mechanisms \cite{MAASS2001ontherelevanceoftime}. 

Temporal coding represents information through the timing of spikes, with several variants. For instance, time-to-first-spike (TTFS) encodes data at the precise time of the first spike, while interspike intervals (time difference between two consecutive spikes) use relative timing to convey information. 
We primarily focus on the TTFS coding since it highlights the characteristics of temporal coding:  efficient representation of information with sparse spiking activity, as the focus on first spikes inherently biases the network toward fewer spikes \cite{gerstner_kistler_naud_paninski_2014}. However, this sparse representation might introduce challenges, including a lack of robustness to noise, as precise spike timing is highly sensitive to perturbations. 
In the remainder, we consider TTFS coding as an affine map, i.e., the encoder $E:\R^{n_{\rm in}}\rightarrow \R^{n_0}$ and decoder $D:\R^{n_{L}}\rightarrow \R^{n_{\rm out}}$ are expressed via parameters $W_{E} \in \R^{n_{0} \times n_{\rm in}}, b_{E} \in \R^{n_{0}}, W_{D} \in \R^{n_{\rm out} \times n_{L}}, b_{D} \in \R^{n_{\rm out}}$ as
\begin{equation}
\label{eqn:encoding_decoding_TTFS}
        E(x) = W_{E}x + b_{E} \text{ for } x \in \R^{n_{\rm in}} \quad \text{ and } \quad D(z) = W_{D}z + b_{D} \quad \text{ for } z \in \R^{n_{L}}.
\end{equation}
Rate coding relies on more robust features such as the frequency and number of spikes at the cost of the additional computational overhead linked to the increased number of spikes to propagate information. For instance, the \emph{spike rate} ($W_D=\text{Id}$, $b_D=0$ in \eqref{eq:rateCod}), which for the discretized LIF model can be coupled with \emph{direct encoding} convenient for static inputs $x\in \R^{n_{\text{in}}}$, generalizes for $s=(s(t))_{t\in[T]}\in \R^{n_L\times T}$ to
\begin{equation}\label{eq:rateCod} 
E(x) = (s^0(t))_{t\in [T]}, \text{where }s^{0}(t)=x\text{ for all }t, \quad \text{ and } \quad D(z)=\frac{1}{T} \sum^T_{t=1} W_D z(t) + b_D.
\end{equation}

\subsubsection{Scope of the survey} \label{sec:Motivation}

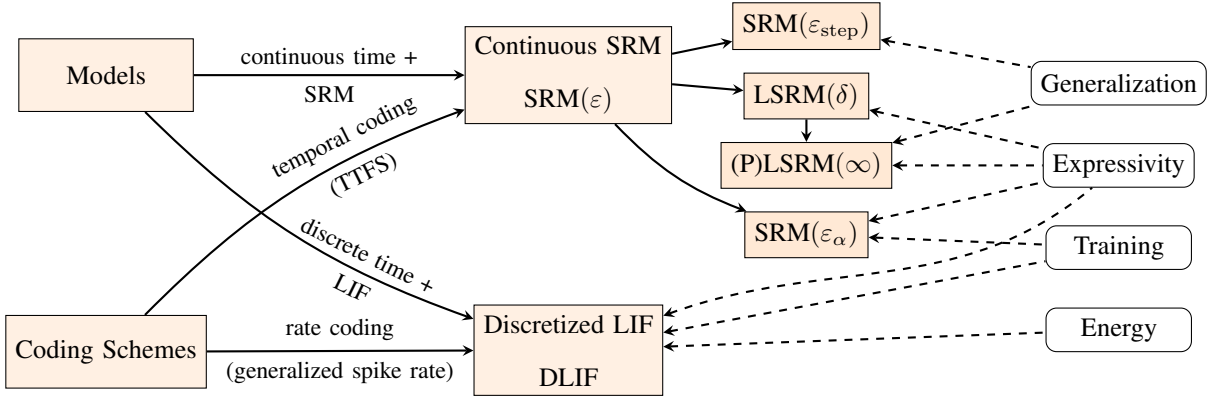
\begin{figure}
    \centering
    \begin{tikzpicture}[scale=0.7]
    \tikzstyle{block} = [rectangle, draw, text centered, minimum height=2.5em, minimum width=6em]
    \tikzstyle{subblock} = [rectangle, draw, text centered, minimum height=1.5em, minimum width=4em]
    \tikzstyle{arrow} = [thick,->,>=stealth]
    \tikzstyle{arrowR} = [thick,<-,>=stealth]

    \definecolor{expressivityColor}{RGB}{30,144,255} 
    \definecolor{trainingColor}{RGB}{220,20,60}      
    \definecolor{generalizationColor}{RGB}{34,139,34} 
    \definecolor{energyColor}{RGB}{148,0,211}        
    \definecolor{Peach}{RGB}{255, 218, 185}
    
    \node [block, fill=Peach!40] (models) {\small Models};
    \node [block, fill=Peach!40, right=3.6cm of models, align=center] (csrm) {\small Continuous SRM \\ \small $\text{SRM}(\varepsilon)$};
    \node [block, fill=Peach!40, below=2.4cm of csrm, align=center] (dlif) {\small Discretized LIF \\ \small DLIF};
    \node [block, fill=Peach!40, below=2.7cm of models] (coding) {\small Coding Schemes}; 
    
    \draw [arrow] (models) -- (csrm) node[above, midway] {\footnotesize{continuous time +}} node[below, midway] {\footnotesize{SRM}};
    \draw [arrow] (coding) -- (dlif) node[above, midway] {\footnotesize{rate coding}} node[below, midway] {\footnotesize{(generalized spike rate)}};
    \draw [arrow] (models)  to [bend right=12.5]  node[sloped, above, pos=0.7] {\footnotesize{discrete time +}} node[sloped, below, pos=0.7] {\footnotesize{LIF}} (dlif);
    
    \draw [arrow] (coding)   to [bend left=12.5] node[sloped, above, pos=0.7] {\footnotesize{temporal coding}}  node[sloped, below, pos=0.7] {\footnotesize{(TTFS)}} (csrm);

    \node [subblock, fill=Peach!60,  right=0.1cm and 0.8cm of csrm.north east] (srm_step) {\small $\text{SRM}({\varepsilon_{\rm step}})$};
    \node [subblock, fill=Peach!60, below=0.3cm of srm_step] (lsrm) {\small $\text{LSRM} (\delta)$};
    \node [subblock, fill=Peach!60, below=0.3cm of lsrm] (plsrm) {\small $\text{(P)LSRM}(\infty)$};
    \node [subblock, fill=Peach!60, below=0.3cm of plsrm]  (srm) {\small $\text{SRM}(\varepsilon_\alpha)$};

    \draw [arrow] (csrm)  to [bend right=12.5] (srm); 
    \draw [arrow] (csrm) -- (lsrm);
    \draw [arrow] (csrm) -- (srm_step);

    \draw [arrow] (lsrm) -- (plsrm);
   
    \node [right=2cm of plsrm, draw, rounded corners, text centered, minimum height=1.5em, minimum width=5em](expr) {\small Expressivity};
    \node [below=0.5cm of expr, draw, rounded corners, text centered, minimum height=1.5em, minimum width=5em](tr) {\small Training};
    \node [above=0.5cm of expr, draw, rounded corners, text centered, minimum height=1.5em, minimum width=5em](gen) {\small Generalization};
    \node [below=0.5cm of tr, draw, rounded corners, text centered, minimum height=1.5em, minimum width=5em] (en) {\small Energy};

    \draw [arrowR,dashed] (srm) -- (expr); 
    \draw [arrowR,dashed] (lsrm) -- (expr);
    \draw [arrowR,dashed] (plsrm) -- (expr);
    \draw [arrowR,dashed] (dlif) to[out=20, in=-140](expr);

    \draw [arrowR,dashed] (srm) -- (tr); 
    \draw [arrowR,dashed] (dlif) -- (tr);
    
    \draw [arrowR,dashed] (plsrm) -- (gen);
    \draw [arrowR,dashed] (srm_step) -- (gen);

    \draw [arrowR,dashed] (dlif) -- (en);

    \end{tikzpicture}
    \caption{A schematic illustrating the design choices in deriving the considered SNN models and highlighting their application in analyzing the expressivity, training, generalization, and energy-efficiency of SNNs.}
    \label{fig:scheme}
\end{figure}

We only introduced a subset, representing the commonly employed approaches in practice, from the wide variety of models and coding schemes. Nevertheless, they allow us to study key properties of SNNs while acknowledging their diversity. To that end, we distinguish SNNs along two primary axes: (i) continuous or discrete handling of the time dimension, and (ii) information propagation through the network either as discrete signals (spike or no spike) or using temporal dimensions (precise timing of spike). Note that both features are critical for SNN performance with certain combinations of models and coding schemes aligning particularly well due to their inherent characteristics.
We analyze the implications of continuous versus discrete time dynamics as well as discrete versus continuous information propagation on the practical and theoretical capacity of SNNs via representatives of the distinct classes: SRM paired with TTFS coding and discretized LIF models paired with spike rate; see \Cref{fig:scheme}. The former combination represents a class of SNNs where both time and information are treated continuously, whereas the underlying process in the latter is discrete in both dimensions---note that the realization might nevertheless map from and to continuous domains taking the coding layers into account. 
To contextualize the findings, we compare these SNN classes to feedforward ANNs, a well-established computational paradigm that typically operates as a discrete-time, continuous information model. Thus, the strengths of the respective classes as well as areas with challenges are identified. 
Thereby, our comparison is grounded on three cornerstones of learning theory: expressivity, which determines whether the model is capable to cope with diverse tasks; training, which assesses whether the model can effectively learn from data to handle practical problems; and generalization, which measures how well it performs on unseen data. Additionally, we incorporate energy efficiency as a fourth dimension, investigating whether the envisioned energy gains of SNNs materialize in computational settings.

\subsection{Statistical learning theory viewpoint}
\label{subsection:stat_learning}
To formally assess model performance, we next introduce key concepts from statistical learning theory  \cite{Anthony_Bartlett_1999}. 
We characterize a learning problem by an \emph{input space} $\gX$, which is assumed to be a compact Euclidean space, a \emph{target space} $\gY \subseteq [-M, M]$, $M \in [0, \infty)$, and a probability distribution $D$ over $\gX \times \gY$.
The goal in statistical learning theory is to select a function $f$ from the space of measurable functions $\gM(\gY^{\gX})$ that best fits $D$ based on some loss function $\mathcal{L}: \gX\times \gY\to \R_{+}$, i.e.,  minimizes the \emph{risk}
\begin{equation*}
    \gE(f) = \E_{X,Y \sim D} \mathcal{L}(f(X),Y).    
\end{equation*}
Typically, we restrict the minimization of the risk to a \emph{hypothesis class} $\gH \subset \gM(\gY^{\gX})$. 
Specifically, we seek the function $f_{\gH} = \text{argmin} \set{\gE(f): f \in \gH}$ 
with the smallest \emph{approximation error} 
\begin{equation*}
    \gE(f_{\gH})-\min_{f\in \gM(\gY^{\gX})} \gE(f)= \min_{f\in \gH}\gE(f)-\min_{f\in \gM(\gY^{\gX})}\gE(f),
\end{equation*}
which can be regarded as a measure for the expressivity of the hypothesis class.
Since the distribution $D$ is unknown, we cannot compute the risk $\gE$ directly. Instead, one aims to minimize the \emph{empirical risk}
\begin{equation*}
    \hat{\gE}(f) = \frac{1}{m}\sum_{i=1}^{m} \mathcal{L}(f(x_i),y_i) \quad \text{ for } m \in \N \text{ i.i.d samples } \{(x_i,y_i)\}^m_{i=1} \sim D^m  
\end{equation*}
to obtain an \emph{empirical risk minimizer}\footnote{ $\min\{\hat{\gE}(f): f\in\gH\}$ is known as the empirical risk minimization (ERM) problem.} $\hat{f}_{\operatorname{erm}} = \text{argmin} \{\hat{\gE}(f): f \in \gH\}$ such that
\begin{equation*}
    \underbrace{\gE(\hat{f}_{\operatorname{erm}})-\min_{f\in\gM(\gY^{\gX})} \gE(f)}_{\text{Excess risk for ERM}}=\underbrace{\gE(\hat{f}_{\operatorname{erm}})-\gE(f_{\gH})}_{\text{Estimation error}}+\underbrace{\gE(f_{\gH})-\min_{f\in\gM(\gY^{\gX})} \gE(f)}_{\text{Approximation error}},
\end{equation*}
where the estimation error can be further decomposed as 
\begin{align*}
    \gE(\hat{f}_{\operatorname{erm}})-\gE(f_{\gH}) &= \underbrace{\gE(\hat{f}_{\operatorname{erm}})-\hat{\gE}(\hat{f}_{\operatorname{erm}})}_{\text{Generalization error for ERM}}+\underbrace{\hat{\gE}(\hat{f}_{\operatorname{erm}})-\hat{\gE}(f_{\gH})}_{\leq 0}+ \underbrace{\hat{\gE}(f_{\gH})-
    \gE(f_{\gH})}_{\text{controlled by classical tools}}  \\
    &\leq \sup_{f \in \gH} \abs{\gE(f) - \hat{\gE}(f)}+ \underbrace{\left|\hat{\gE}(\hat{f}_{\gH})-\min_{f\in\gH} \gE(f) \right|}_{O(\sqrt{1/m}) \text{ with high probability}}.
\end{align*}
Hence, the goal of bounding the excess risk for the ERM can be accomplished by bounding the \emph{approximation error}, which leads to the study of expressivity, and the \emph{generalization error}, which can be bounded by $\sup_{f\in \gH}|\gE(f) - \hat{\gE}(f)|$. Thereby, the latter quantifies the gap between the expected and empirical risk, providing theoretical guarantees for a model’s learning ability. Arguably, the main pitfall of this (uniform bound) approach is that it is training independent. As can be seen from the previous decomposition, one avenue for improvement comes from a consideration of the \emph{training error} $\hat{\gE}(\hat{f}_{\operatorname{erm}})$.


\begin{figure}
    \centering
    \begin{subfigure}{0.4\textwidth}
        \centering
        \begin{tikzpicture}[scale=0.7]
            \begin{axis}[
            ylabel={$\varepsilon_\alpha(t)$},
            xlabel={$t$},
                domain=0:8, samples=200,
                axis lines=middle, 
                enlargelimits=false, clip=false,
                ymin=0,ymax=1, legend cell align={left},
                axis line style={thick, -{Stealth[scale=1.5]}}
            ]
                \addplot[blue, thick] {x*exp(-x*0.5)}; 
                \addlegendentry{$\tau_c=0.5$}
                \addplot[green, thick] {x*exp(-x*1)};  
                \addlegendentry{$\tau_c=1$}
                \addplot[green, thick, dashed] {0.5*x*exp(-x)}; 
                \addlegendentry{$\tau_c=1, w=0.5$}
                \addplot[green, thick, dotted] {2*x*exp(-x)}; 
                \addlegendentry{$\tau_c=1, w=2$}
                \addplot[red, thick] {x*exp(-x*2)};  
                \addlegendentry{$\tau_c=2$}

                \node[black, right] at (axis cs:4,0.2) {\footnotesize{approx. linear part}};
                \draw[->,>=stealth,dashed, black] (axis cs:5,0.25) -- (axis cs:0.8,0.4);
                \draw[->,>=stealth,dashed, black] (axis cs:5,0.25) -- (axis cs:3.95,0.5);
                
            \end{axis}
        \end{tikzpicture}
        \caption{\phantom{x}}
        \label{fig:response_a}
    \end{subfigure}
    \hspace{1.5cm}
    \begin{subfigure}{0.4\textwidth}
        \centering
        \begin{tikzpicture}[scale=0.7]
            \begin{axis}[
            xlabel={$t$},
            ylabel={$\varepsilon_y(t)$},
            ylabel style={rotate=-90}, 
                domain=-0.5:1.25, samples=100,
                axis lines=middle, 
                enlargelimits=false, clip=false,
                ymin=0, ymax=2,xmin=-0.5,xmax=1.5,  legend cell align={left},
                axis line style={thick, -{Stealth[scale=1.5]}}
            ]

                \addplot[red, thick,domain=0:0.001] {0.01};
                \addlegendentry{lin. $\delta=0.5$}
                \addplot[red, thick, dashed, domain=0:0.001] {0};
                \addlegendentry{lin. $\delta=\infty$};
                \addplot[blue, thick,domain=0:0.001] {0};
                \addlegendentry{step};
                
                \addplot[red, thick,domain=-1:0] {0.01};
                \addplot[red, thick, domain=0:0.5] {x};
                \addplot[red, thick, domain=0.5:1.5] {0.01}; 
                \addplot[red, thick, dashed, domain=0.5:1.5] {x}; 
                \addplot[red, thick, ycomb] coordinates {(0.5,0.5)};
                \addplot[red, thick, domain=0.5:1.5] {0.01}; 
                
                \addplot[blue, thick, domain=-1:0] {0};
                \addplot[blue, thick, ycomb] coordinates {(0,1)};
                \addplot[blue, thick, domain=0:1] {1};
                \addplot[blue, thick, domain=1:1.5] {0};
                \addplot[blue, thick, ycomb] coordinates {(1,1)};
                               
            \end{axis}
        \end{tikzpicture}
        \caption{\phantom{x}}
        \label{fig:response_b}
    \end{subfigure}
    \caption{Illustration of different response functions in the SRM model. (a) Biologically realistic response function $\varepsilon_\alpha$ with parameter $\tau_c$ and weights $w$. (b) Simplified response functions abstracted from $\varepsilon_\alpha$, including a piecewise linear function with cutoff at $\delta=0.5$, a step function, and the ReLU activation ($\delta=\infty$). The delay as an additional parameter allows for shifted versions along the time dimension.}
    \label{fig:response}
\end{figure}
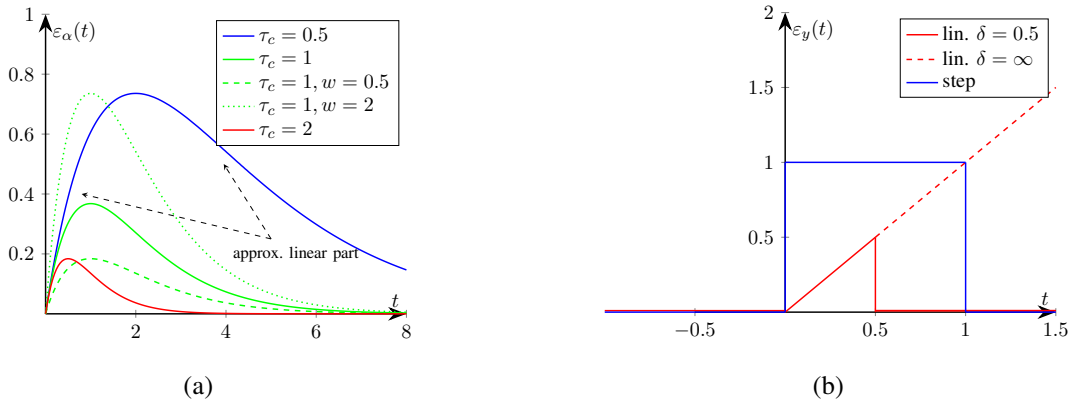

\section{Expressivity}
\label{subsection:expressivity}
The spike-based communication of SNNs alongside their complex dynamics makes the structure and organization of computations challenging to analyze. 
Nevertheless, we begin by emphasizing the similarities between the expressive power of ANNs and SNNs in continuous time using temporal coding to establish a shared foundational understanding before examining the structural differences in the computational frameworks. To do so, we focus on the SRM model with TTFS coding assuming additionally that each neuron spikes only once.  
The restriction to one spike is not a general condition but reflects a model's tendency towards sparse spiking in TTFS and entails a simplification in the computation of the firing time via \eqref{eq:SRMST}. Since multiple spikes (and thereby refractoriness effects) are now disregarded, the time-dependent threshold $\vartheta$ can be treated as a constant so that the firing time $t_i^f$ of a neuron $i$ can be computed via
\begin{equation}\label{eq:TTFSFiringTime}
    t_i^f = \min\{t: t > \min_j t_j^f \text{ and } u_i(t) = \sum_{j} w_{i,j}\varepsilon(t-(t_j^f + d_{i,j})) = \vartheta\}.
\end{equation}
We denote the resulting model class mainly dependent on the response kernel $\varepsilon$ as
\begin{equation*}
    \gS_{\text{SRM}}(\varepsilon) := \big\{\Psi: \Psi =(\Phi, (E,D)) \text{ with } \Phi = ((W^\ell, D^\ell, \vartheta^\ell)_{\ell \in L})  \text{ and } E, D \text{ given in } \eqref{eqn:encoding_decoding_TTFS} \big\}.   
\end{equation*} 
However, deriving an analytical solution for the firing time still requires $\varepsilon$ to have some favorable properties. 
A biologically realistic response kernel closely modeling the shape of synaptic responses in neurons is given by $\varepsilon_\alpha(t) = t\exp(-t\tau_c)$, $\tau_c>0$; see \Cref{fig:response_a}. In this context, it was shown in \cite{temporal_single_spike_backprop2020_comsa} that the class $\gS_{\text{SRM}}(\varepsilon_\alpha)$ is a universal approximator of continuous functions under certain conditions. A more thorough analysis of this model class is challenging due to the intricacies of the response $\varepsilon_\alpha$.
To circumvent this issue, a sufficiently simple response kernel as in the \emph{linearized SRM} (LSRM) can be considered, i.e.,
\begin{equation} \label{eqn:Linear_SRM} 
    \gS_{\text{LSRM}}(\delta) := \gS_{\text{SRM}}(\varepsilon^\delta_{\text{lin}}) \text{ with } \varepsilon^\delta_{\text{lin}}(t) =  
    \begin{cases}
        0, & \text{if } t  \notin [0, \delta]\\
        t ,  & \text{if } t \in [0, \delta]
    \end{cases} 
    \quad \text{ for } \delta >0.
\end{equation} 
Thus, $\varepsilon^\delta_{\text{lin}}$ consists of a linear segment of length $\delta$, only roughly resembling a biological response. The intuition behind the model is to zoom in on a linear segment of more complex response functions and neglect other aspects by postulating that all incoming spikes arrive in this time window; see \Cref{fig:response_b}. 
In this setting, the first universal approximation result for SNNs was given by Maass \cite{Noisyspikingneurons_Maass1996} by reducing the problem to ANNs with (continuous) piecewise linear activation function \cite{leshno1993_ridge}. 

\begin{theorem}
\label{thm:maassUAT}
Let $f: [0,1]^d \rightarrow [0,1]$ be continuous. For all $0<\epsilon<1$, there exists for all $\delta\geq3$ an SNN $\Psi \in \gS_{\text{LSRM}}(\delta)$ with $L=1$ such that $R(\Psi)$ uniformly approximates $f$ with $\epsilon$-accuracy.
\end{theorem}
The universal approximator property can be straightforwardly extended to arbitrary depths as well as explicit approximation rates introduced by a thorough analysis of the constructive proof. 
The recent result \cite{STANOJEVIC2023}, though aimed at training, presents another universality result along with complexity bounds in a specific IF model under TTFS encoding. 
They derive a neuron-to-neuron mapping that converts ReLU-based ANN parameters to SNNs while preserving function realization, enabling the transfer of known ANN approximation rates to the spiking domain. However, given their distinct internal mechanisms, can we demonstrate meaningful differences in the capabilities of ANNs and SNNs?

In this direction, Maass already observed that SNNs exhibit a superior capacity for specific (biologically relevant) toy problems \cite{Maass1996Networksthirdgeneration}. 
To further explore the distinctions between SNNs and ANNs, we will focus on a special class of $\gS_{\text{LSRM}}(\delta)$ by setting $\delta=\infty$, i.e., the corresponding response function $\varepsilon^\infty_{\text{lin}}(t)=\max\{0,t\}$ is simply the ReLU activation (\Cref{fig:response_b}). 
Note that a large linear segment has a constant effect on postsynaptic neurons, meaning spikes lose their point-like temporal nature. In contrast, smaller, biologically realistic linear segments require tightly synchronized spike timings to jointly affect a neuron's potential, as earlier spikes may decay before later ones contribute. From a computational perspective, this implies that small $\delta$ leads to additional complexity since the same firing patterns yield different outcomes. 

Applying the response kernel $\varepsilon^\infty_{\text{lin}}$ together with the restriction to positive weights leads to the class $\gS_{\text{PLSRM}}=\{\Psi\in \gS_{\text{LSRM}}(\infty): (W^\ell)_{\ell \in L}\geq0\}$ of \emph{positive, linearized SRMs} (PLSRM). These were shown to be universal approximators in \cite{NP2024expressivitySNN} by approximating certain ridge functions and leveraging the universality of finite sums of ridge functions \cite{leshno1993_ridge}.
A key step therein as well as in achieving dimension-independent approximation rates for Barron-regular functions and optimal rates for smooth functions is the approximation of the minimum function on $\R^d$ by an SNN $\Psi \in \gS_{\text{PLSRM}}$ consisting of a single spiking neuron with $d$ input neurons. This indicates the potential for effectively approximating linear finite element spaces composed of piecewise affine functions \cite{NP2024expressivitySNN} and highlights a difference to ReLU-ANNs, which, regardless of depth, must have at least $d$ neurons per hidden layer or at least depth three to efficiently approximate the minimum function under certain conditions \cite{SafranRV24}. However, in contrast to ReLU-ANNs, which can realize any continuous piecewise linear function (CPWL), PLSRMs cannot realize all CPWL functions or even efficiently approximate certain CPWL functions such as sawtooth functions.

Re-introducing negative weights in the PLSRM class enables the realization of (jump-)discontinuous mappings \cite{SFK2024expressivitySNN}, which represents a distinction between LSRMs and ReLU-ANNs. 
However, by ensuring that each neuron’s incoming weights sum positively and setting sufficiently high thresholds, the realization of LSRMs is CPWL. Under these conditions, the class $\gS_{\text{LSRM}}(\infty)$ emulates any ReLU-ANN and thus realizes any CPWL function as well as reproduces all approximation results by ReLU-ANNs with similar approximation rates, addressing some of the limitations of PLSRMs \cite{SFK2024expressivitySNN}. 
However, further exploration is needed to investigate the fine differences in the structure of computations between SNNs and ANNs. 
Some preliminary results revealing deviations, particularly in the scaling behavior of the number of linear regions, were presented in \cite{SFK2024expressivitySNN}. 
Finally, we note that some results exist in the continuous-time setting based on rate coding. In \cite{zhang2024intrinsicSNN, firingrateExpressivity_zhang2022}, it is shown that self-connected SNNs approximate continuous functions, radial functions, and dynamical systems with polynomial complexity in both parameters and time.

Switching from the continuous time to the discrete time framework entails a loss of information. Moreover, assuming binary information propagation (in contrast to the previously considered cases) as in the discretized LIF model further contributes to this effect. Hence, an immediate question is whether networks of discretized LIF neurons retain the same expressive power as the continuous time models when operating on the same domain. Formally, based on \eqref{eq:LIFSNN}, the class of discretized LIF (DLIF) SNNs 
\begin{equation*}
    \gS_{\text{DLIF}} := \big\{\Psi: \Psi =(\Phi, (E,D)) \text{ with } \Phi = ((W^\ell, b^\ell, u^\ell(0), \beta^\ell, \vartheta^\ell))_{\ell \in L}, T)  \text{ and } E, D \text{ given in } \eqref{eq:rateCod} \big\},   
\end{equation*} 
realizes boolean functions when neglecting the coding layers. Moreover, enrolling the time dimension in the model dynamics \eqref{eq:LIFSNN}, the realization of $\Psi \in \gS_{\text{DLIF}}$ composes the Heaviside and affine functions 
\begin{align}\label{eq:HANN}
    R(\Psi) &= D \circ H \circ A^L(\cdot) \circ H \circ  \hdots \circ A^2(\cdot) \circ H \circ A^1(\cdot) \circ E \quad \text{ with } A^\ell(\cdot): \R^{n_{\ell-1} \times T}\to \R^{n_\ell \times T}\\
    &\text{ given by } \big(A^\ell(\tilde{b})\big)(s) =  W^\ell  s +\tilde{b} \quad \text{ for } s=(s(t))_{t\in [T]} \in  \R^{n_{\ell-1} \times T}, \tilde{b}=(\tilde{b}(t))_{t\in [T]} \in  \R^{n_{\ell} \times T}. \nonumber
\end{align}
Note that the specific form of $A^\ell(\cdot)$ depends on the variable $\tilde{b}$, which represents the dynamical aspects including the evolution of the potential of the neuron. In other words, the dynamics (and thereby a chunk of the computational overhead) are outsourced and hidden in the input-dependent variable $\tilde{b}$. However, from a theoretical perspective, \eqref{eq:HANN} highlights a link to feedforward ANNs. For $T=1$, the model is equivalent to ANNs with Heaviside activation function since there are no temporal dynamics, which need to be taken into account so that the layer-wise affine mapping is simply $A^\ell(s) = W^\ell s + \tilde{b}$ for some fixed $\tilde{b}\in \R^{n_{\ell}}$. Hence, the universal approximation properties of Heaviside ANNs with respect to Boolean as well as continuous functions extend also to the class $\gS_{\text{DLIF}}$. For $T>1$, the equivalence between the ANN and $\gS_{\text{DLIF}}$ is not valid anymore, however, structural similarities remain and can potentially be exploited, e.g., the universal approximation property can easily be extended to this case as well.

\section{Training}
\label{section:training}

The training process involves selecting an SNN that best aligns with the given training data according to a specified criterion. Typically, it consists in finding the empirical risk minimizer $\hat{f}_{\text{erm}}$ defined in Section \ref{subsection:stat_learning}, 
where a specific class of SNN is selected as the hypothesis set $\gH$.
The primary challenges in solving this optimization problem, shared by both SNNs and ANNs, are its inherent high dimensionality (due to the large number of parameters) and the non-convexity of the empirical risk generally making the problem NP-hard. Despite these difficulties, first-order methods such as gradient descent and its variants---including stochastic gradient descent (SGD)---have proven highly effective in practice for ANNs, even achieving remarkable generalization performance, by using gradient information to control the training dynamics. These methods have become the gold standard for training ANNs with ReLU and differentiable activations. 

SNNs can be categorized as either differentiable or non-differentiable, depending on whether their output is differentiable with respect to their parameters. 
In the differentiable case, such as the $\gS_{\text{SRM}}(\varepsilon)$ class, a straightforward implementation of gradient descent is possible despite some obstacles unique to the spiking regime. Conversely, for non-differentiable models, like the $\gS_{\text{DLIF}}$ class, different approaches are required. Several strategies have been proposed, including: (i) \emph{ANN to SNN conversion}, which bypasses the problem by training an ANN and then converting it to an SNN; (ii) \emph{Direct training with surrogate gradients}, which replaces non-differentiable activations with smooth approximations; and (iii) \emph{Local learning rules}, which involve biologically inspired weight update mechanisms such as spike-timing dependent plasticity (STDP).

\subsection{Training in the Discretized LIF model}
\label{subsection:training_surrogate}

For the $\gS_{\text{DLIF}}$ class, we focus on training using the surrogate gradient approach, which is arguably the most popular method for direct training \cite{SNNSurvey2023}.
This technique was originally developed to address the challenge of training ANNs with Heaviside activations. It builds on the \emph{backpropagation through time} (BPTT) algorithm for gradient computation but introduces a key modification: during the backward pass, the non-differentiable activations are replaced with differentiable surrogates, enabling the calculation of gradients. Importantly, the original non-differentiable activations are retained during the forward pass \cite{Neftci_SurrogateGD_SNNs_2019}.

\subsubsection{Learning setup} For clarity, we consider supervised learning with `static' labels, under the following training setup. Given time-dependent data $(x[k])_{k \in [m]}$
and corresponding labels 
\(
(y[k])_{k \in [m]},
\)
the objective is to find $\Psi = (\Phi,(E,D)) \in\gS_{\text{DLIF}}$ that minimizes the empirical risk for a given loss function 
\(
\mathcal{L} : \mathbb{R}^{n_{\text{out}}} \times \mathbb{R}^{n_{\text{out}}} \to \mathbb{R}
\), whose choice is determined by the practitioner based on the specific task at hand. 
Assuming that the learnable parameters are encoded in $\Phi$ (which can generally be extended to learnable $E, D$ as well), the optimization problem we aim to solve can now be expressed as 
\begin{align}\label{eq:assumption_predictor}
    \min_{\Phi} \hat{\mathcal{E}}(R(\Phi,E,D)) &=\min_{\Phi}\frac{1}{m}\sum_{k=1}^{m} \gL\Big(R(\Phi,E,D)\big(x[k]\big),y[k]\Big)\\
    &= \min_{\Phi}\frac{1}{m}\sum_{k=1}^{m} \gL\Big(\frac1T\sum^T_{t=1}s^L[k](t),y[k]\Big) \text{ with } (s^L[k](t))_{t\in[T]} = (R(\Phi) \circ E)(x[k]). \notag
\end{align}
To employ gradient-based methods such as SGD, the crucial step is the calculation of the gradient of $\gL$.

\subsubsection{Backpropagation through time (BPTT)}

Implementing backpropagation in this context requires accounting for the temporal dimension when applying the chain rule. 
The gradient of $\gL$ has the form
\[
\nabla_{\Phi}\gL=\left(\frac{\partial\gL}{\partial W^\ell_{ij}}, \frac{\partial\gL}{\partial b^\ell_{i}},\frac{\partial\gL}{\partial \beta^\ell},\frac{\partial\gL}{\partial \vartheta^\ell}\right)_{i\in [n_{\ell-1}],j\in[n_\ell],\ell\in[L]},
\]
which is composed by the derivatives with respect to each of the trainable parameters. 
%
%
Using the chain rule, the derivatives of $\gL$ with respect to the elements in $W^\ell$ can be formally expressed as
\begin{equation}\label{eq:chain_rule1}
  \frac{\partial\gL}{\partial W^\ell_{ij}} = 
\sum^T_{t=1}\sum^{n_L}_{j'=1} 
\frac{\partial\gL}{\partial s^L_{j'}(t)} 
\frac{\partial s^L_{j'}(t)}{\partial u^L_{j'}(t)}
\frac{\partial u^L_{j'}(t)}{\partial W^\ell_{ij}} \quad \text{ for any } i\in [n_{\ell-1}], j\in[n_\ell].  
\end{equation}
The issue here is the non-differentiability of the (binary) spikes $\{s^L_{j'}\}_{j'\in [n_L]}$, which makes the term $\frac{\partial s^L_{j'}}{\partial u^L_{j'}}$, in \eqref{eq:chain_rule1}, not well-defined. Replacing the Heaviside activation with a differentiable surrogate $h_{\text{sg}}$, we obtain
\begin{equation*}
    \frac{\partial s^L_{j'}(t)}{\partial u^L_{j'}(t)} = h'_{\text{sg}}\left(u^L_{j'}(t)-\vartheta^L\right) \quad \text{and} \quad \frac{\partial \gL}{\partial s^L_{j'}(t)}=\frac1T \frac{\partial \gL}{\partial x} s^L_{j'}.
\end{equation*}
Similarly, using the model dynamics \ref{eq:LIFSNN}, we verify, for $\ell\leq \ell'\leq L $ and $j'\in [n_{\ell'}]$
\[
\psi_{i,j,\ell}(\ell',t;j'):=\frac{\partial u^{\ell'}_{j'}(t)}{\partial W^\ell_{ij}}=\beta \frac{\partial  u^{\ell'}_{j'}(t-1)}{\partial W^\ell_{ij}} + 
\sum^{n_{\ell'-1}}_{i'=1}W^{\ell'-1}_{i'j'} \frac{\partial  s^{\ell'-1}_{j'}(t)}{\partial u^{\ell'-1}_{j'}(t)} \frac{\partial u^{\ell'-1}_{i'}(t)}{\partial W^\ell_{ij}}
\]
to obtain with $\frac{\partial  s^{\ell'-1}_{j'}(t)}{\partial u^{\ell'-1}_{j'}(t)} = h_{\text{sg}}'\left(u^{\ell'-1}_{i'}(t)-\vartheta^{\ell'-1}\right)$ the recurrence
\begin{equation}\label{eq:recurrence}
    \psi_{i,j,\ell}(\ell',t;j')=\beta \psi_{i,j,\ell}(\ell',t-1;j')+ \sum^{n_{\ell'-1}}_{i'=1}W^{\ell'-1}_{i'j'} h_{\text{sg}}'\left(u^{\ell'-1}_{i'}(t)-\vartheta^{\ell'-1}\right)\psi_{i,j,\ell}(\ell'-1,t;i').
\end{equation}
The first term on the right-hand side of \eqref{eq:recurrence} represents a temporal recurrence, while the second term corresponds to a recurrence across layers. The base elements of this recurrence are the terms $\{u^{\ell'}(0)\}_{\ell \leq \ell' \leq L}$ and $\{s^{\ell-1}(t)\}_{t \in [T]}$, where the former serves as a model hyperparameter, and the latter represents the input to the sub-network starting from layer $\ell$.
With this, the recurrence in \eqref{eq:recurrence} can be solved, which completes the calculation of $\frac{\partial\gL}{\partial W^\ell_{ij}}$.
The expressions for the derivatives of $\gL$ with respect to $b^\ell, \beta^\ell$ and $\vartheta^\ell$ are analogous. It is important to highlight that most terms in the preceding calculations are independent of the choice of the loss function. The only exception is the term $ \frac{\partial \gL}{\partial s^L_{j'}(t)}$, which depends on the loss function exclusively through $ \frac{\partial \gL}{\partial x} $.
Several loss functions have been proposed for supervised tasks with SNNs often analogous to their counterparts in ANNs but adapted to account for the specific characteristics of SNN predictors.  
For instance, in \emph{classification} tasks, common objectives include the cross-entropy loss and mean squared error loss \cite[Appendix B]{SNNSurvey2023}. 
While the selection of appropriate surrogate functions has largely been driven by empirical studies, a more rigorous theoretical analysis is warranted. The sigmoid and arctangent functions are frequently employed as surrogates, with the latter demonstrating superior performance across multiple application domains \cite{fang2021deep}. Utilizing regularization methods like the $\ell_1$-norm have also been explored \cite{SNNSurvey2023}, however, they require careful implementation. Excessive penalization of spike counts may suppress neuronal activity, potentially leading to convergence issues and training stagnation.

\subsection{Training in continuous-time SRM model with TTFS encoding}
\label{subsubsection:training_TTFS}

In the  $\gS_{\text{SRM}}(\varepsilon)$ class the differentiable relationships are shifted into the temporal domain by leveraging the precise timing of individual spikes. Therefore, the backpropagation strategy involves finding the differentiable relationship of the postsynaptic spike time with respect to synaptic weights and presynaptic spike times (neglecting the synaptic delays for clarity). While the original idea stems from the SpikeProp model \cite{spikeprop}, we focus on the extension in \cite{temporal_single_spike_backprop2020_comsa}; a similar approach for LIF neurons is covered in \cite{firstspike2021goltz}. 

\subsubsection{Learning setup}
We again focus on a supervised learning approach, more exactly on a classification task with $c$ classes, i.e., the data consists of inputs $(x[k])_{k\in[m]}$ with corresponding one-hot encoded labels $(y[k])_{k\in[m]} \in \{0,1\}^{c \times m}$ and the hypothesis class is $\{\Psi \in \gS_{\text{SRM}}(\varepsilon_\alpha): \Psi=(\Phi, \text{Id}, \text{Id}) \text{ with } n_L=c\}$, 
where we assumed for clarity that the coding functions are identity functions on the respective domains. Hence, the firing times of neurons in the first layer are given by $x[k]$ and the prediction of the network corresponds to the index of the neuron in the final layer that spikes first. The learning objective is to optimize spike times via the learnable parameters such that the target neuron fires earlier than non-target neurons, which can be formalized via the softmax function and the cross-entropy loss $\mathcal{L}^{CE}: \mathbb{R}^{n_{\text{out}}} \times \mathbb{R}^{n_{\text{out}}} \to \mathbb{R}$:
\begin{equation*}
    \min_{\Phi}\frac{1}{m}\sum_{k=1}^{m} \gL^{CE}\Big(R(\Phi,\text{Id},\text{Id})\big(x[k]\big),y[k]\Big)
    = \min_{\Phi} - \frac{1}{m}\sum_{k=1}^{m} \sum_{i=1}^c y_i[k]\log{\frac{\exp{(-t_{L,i}^f[k])}}{\sum_{j=1}^{c}\exp{(-t_{L,j}^f[k])}}},
\end{equation*}
where $t_{L,i}^f[k]$ denotes the firing time of neuron $i$ in layer $L$ on input of $x[k]$.

\subsubsection{Backpropagation}

Unlike the discretized case, this model allows for the exact computation of gradients with respect to both spike times and learnable parameters, i.e., weights. We skip the details of backpropagation, which follows the approach described in the previous case, and only highlight the key step in computing the relevant gradients. 
In particular, for a neuron $i$ the firing time $t_i^f$ can be calculated by identifying the minimal subset $J$ of all presynaptic neurons, which cause with their incoming spikes the potential $u_i$ to reach the threshold $\vartheta_i$ while rising, and the formula
\begin{equation*}
    t^f_{i} = \frac{B_J}{A_J} - \frac{1}{\tau_c}W\bigg(\tau_c \frac{\vartheta_i}{A_J}\exp(\tau_c \frac{B_J}{A_J})\bigg), \text{ where } A_J = \sum_{j\in J}w_{i,j}\exp{(\tau_c t^f_j)}, B_{J} = \sum_{j\in J}w_{i,j}\exp{(\tau_c  t_j^f)}t^f_j,  
\end{equation*}
and $W$ denoting the Lambert W function \cite{Corless1996OnTL}. For details concerning the approach to determine the subset $J$, we refer to \cite{temporal_single_spike_backprop2020_comsa}. Instead, we highlight that with the derived description the relevant gradients for the backpropagation process---$\frac{\partial t^f_i}{\partial t^f_j}$ and $\frac{\partial t^f_i}{\partial w_{i,j}}$ for any neuron $j\in J$--- can be analytically derived by exploiting the properties of the Lambert W function (where the restriction to single spikes per neuron turns out to be crucial). 
In practice, slight adaptations increase the performance and address minor issues like vanishing gradients. Moreover, non-differentiable points in the loss function corresponding to a change in the set $J$ do not degrade the performance (similar to ReLU in ANNs). Overall, the computational complexity of this algorithm is a concern, as sorting spikes and solving the Lambert W function can be resource-intensive. However, the model's temporal nature and analytical spike computation align well with neuromorphic hardware implementations, which might ameliorate the aforementioned computational issues. While effective for classification, its suitability for regression tasks remains underexplored.

\subsection{Comparison}
The discretized LIF model, with its ability to generate multiple spikes, allows for richer information and adapts more effectively to sequential data compared to the single-spike constraint of the TTFS framework in the continuous dynamics. However, the continuous-time model inherently leverages the temporal component, enabling fine-grained adjustments based on precise spike timing, which can be advantageous in tasks requiring temporal sensitivity. Despite non-differentiability challenges, the surrogate gradient approach has gained popularity due to its match with current hardware technology and training pipelines, contrasted by recent advances in the continuous time framework with TTFS coding signaling growing interest and progress. Going beyond practical implementations, we are unaware of any work on the theoretical side that analyzes the loss landscape of SNNs with either model.

\section{Generalization}
\label{section:generalization}
The question of why modern, highly overparameterized ANNs perform well on unseen data---often referred to as the \emph{generalization puzzle}---remains one of the key open problems in neural network theory. 
While extensive theoretical and empirical research has focused on understanding generalization in ANNs, results for SNNs remain sparse. The few notable contributions exploring this issue have largely followed a similar trajectory to ANN theory, beginning with techniques rooted in classical learning theory. The \emph{VC-dimension} (and its generalization to real-valued functions, the \emph{pseudodimension}) as well as covering numbers are well-established complexity measures that can be used to bound the generalization error \cite{Anthony_Bartlett_1999}. 

Some of the earliest theoretical results on generalization in SNNs were introduced in \cite{VCdimension_maass1999, Schmitt99VCdim} by specifying their VC-dimension and pseudodimension. By combining a VC-dimension generalization error bound \cite{Anthony_Bartlett_1999, Berner_2022} with the obtained VC-dimension bound $\mathcal{O}(ML\log(ML))$ for any $\Psi \in \gS_{\text{SRM}}(\varepsilon_{\text{step}})$,  $\varepsilon_{\text{step}}(t) = H(t) - H(t+1)$ (\Cref{fig:response_b}), with $M$ edges and depth $L$ the following result is established.

\begin{theorem}
\label{theorem:gen_VC_schmitt} 
    Denote by $\gS^{M,L}_{\text{SRM}}(\varepsilon_{\text{step}}) \subset \gS_{\text{SRM}}(\varepsilon_{\text{step}})$ the class of SRMs with $M$ connections and depth $L$. For every $\delta \in (0,1)$ and data set of size $m \in \N$ sampled according to a distribution $D$ on $\gX \times \{-1,1\}$,  
    we have with probability $1-\delta$
    \begin{equation*}
        \sup_{\Psi \in \gS^{M,L}_{\text{SRM}}(\varepsilon_{\text{step}})} \abs{\gE(R(\Psi)) - \hat{\gE}(R(\Psi))} \leq c\sqrt{\frac{ML\log(ML)+\log (1/\delta)}{m}} \quad \text{ for some } c \in (0, \infty).
    \end{equation*}   
\end{theorem}

Recently, \cite{NP2024expressivitySNN} established generalization bounds for PLSRMs using covering number techniques to control the model complexity. This is based on a boundedness assumption on the hypothesis class of PLSRMs. 
In this result, the relevant covering number bound scales linearly in $M$ and logarithmically with $L$, improving on the above VC-dimension based and classical ANN generalization bounds.%

Finally, in a stochastic variant of the LIF model, where the firing probability of a neuron depends on its potential, \cite{zhang2024intrinsicSNN} provided learning guarantees by bounding the Rademacher complexity, a key measure of a model's capacity. In particular, it is shown that stochasticity improves performance over non-stochastic SNNs and even surpasses the results in ANNs using similar methods. 
However, the theoretical development of ANNs has progressed significantly further. 
For instance, some of the most successful results in ANNs have been achieved through PAC-Bayes theory \cite{dziugaite2017computing}, a framework that remains largely underexplored in the context of SNNs. Therefore, it is crucial to understand whether these more recent advances can be adapted to SNNs and whether the inherent sparsity of SNNs could lead to improved results. Moreover, in overparameterized networks, as noted for ANNs, generalization is influenced by implicit biases introduced during training, which classical learning theory cannot fully explain. Exploring more modern approaches attempting to address this limitation like loss landscape sharpness and phenomena such as double descent in the context of SNNs remains an open question \cite{jiang_gen_measures2020}.

\section{Energy-efficiency and related concepts}
\label{sec:energy}
The interplay of hardware and software determines the potential 
for leveraging spiking-based computations. 
For instance, the performance of training methods may 
depend on the specific hardware used. Currently, the best results in training on classical digital hardware are achieved by adopting ANN training techniques in contrast to incorporating more biologically plausible training mechanisms such as STDP \cite{gerstner_kistler_naud_paninski_2014}. 
However, this observation might be invalidated for other types of hardware such as neuromorphic, which by design supports spike-based processing \cite{Mehonic2024NeuromorphicRoadmap}.
Hence, hardware-software co-design is crucially important for the future impact of SNNs. Next, we will support this claim by analyzing the energy efficiency of SNNs, a key motivation for their practical implementation. Initially, the energy consumption of SNNs relative to ANNs was assessed based on the dynamic energy usage of arithmetic operations. This approach was chosen because the number of operations serves as a hardware-agnostic and easily trackable measure. 
In the remainder of this section, our goal is to highlight the limitations of this measure and present extensions that better align with experimental observations. Thereby, we treat discretized models, since most theoretical work is conducted in this setting for reasons that will become clear shortly. 

The computational operations of a neuron $i$ in a feedforward ANN leading to its output $y_i\in \R$ is
\begin{equation}\label{eq:ANNneuron}
    y_i = \varphi(\sum_j x_j w_{i,j} + b_i), \quad w_{i,j}, b_i \in \R, \quad \varphi:\R\to\R,
\end{equation}
where $x_j\in\R$ is the input from presynaptic neuron $j$, which corresponds to \emph{multiply-and-accumulate} (MAC) operations. In contrast, the computational operations of a neuron in the $\gS_{\text{DLIF}}$ class simplifies to
\begin{equation}\label{eq:LIF_Energy}
    u_i(t_{n+1}) = \beta u_i(t_n) + \sum_j w_{i,j} s_j(t_{n+1}),
\end{equation}
when neglecting the reset mechanism in \eqref{eq:LIFDisFinal} for clarity. This corresponds to \emph{accumulate} (AC) operations, i.e., additions of $w_{i,j}$ to the potential given an associated spike $s_j(t_{n+1}) =1$.
Hence, low-energy AC operations replace energy-intensive MAC operations predicting energy gains by SNNs, assuming the same number of operations with the exact energy difference between MACs and ACs depending on data precision, semiconductor and processor technology, etc. \cite{Horowitz14}.

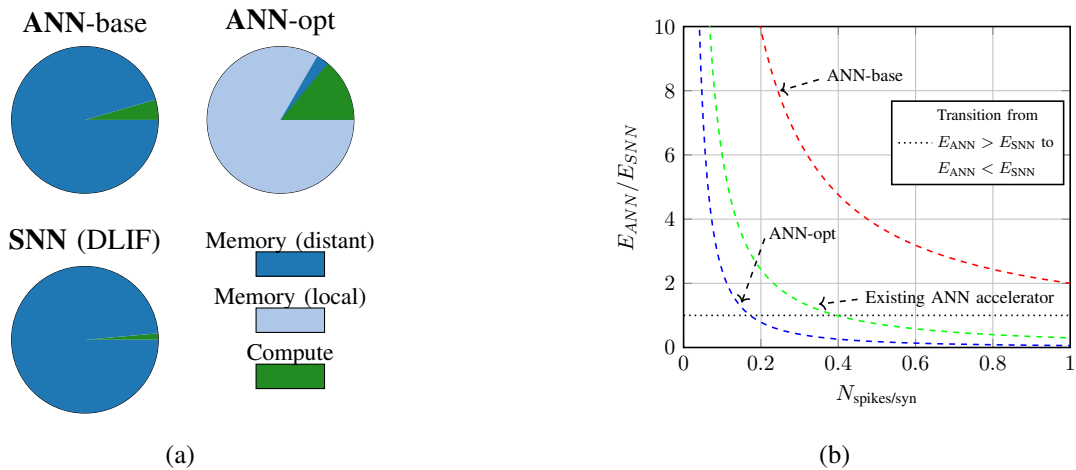
\begin{figure}
    \centering
    \begin{subfigure}{0.45\textwidth}
        \centering
        \begin{tikzpicture}[scale=0.65,rotate=-90]
            \definecolor{memoryColor}{RGB}{31, 119, 180}  
            \definecolor{locmemoryColor}{RGB}{174, 199, 232}   
            \definecolor{computeColor}{RGB}{34,139,34} 
            
            \begin{scope}[shift={(-2.5,-2)}]
                \node at (-2,0) {\textbf{ANN}-base};  
                \draw (0,0) circle (1.5);  
                \fill[computeColor] (0,1.5) arc[start angle=90, end angle=106, radius=1.5] -- (0,0) -- cycle;  
                \fill[memoryColor] (0,0) -- (0,1.5) arc[start angle=90, end angle=-254, radius=1.5] -- cycle;  
            \end{scope}

            \begin{scope}[shift={(-2.5,2)}]
                \node at (-2,0) {\textbf{ANN}-opt};  
                \draw (0,0) circle (1.5);  
                \fill[locmemoryColor] (0,0) -- (0,1.5) arc[start angle=90, end angle=-215, radius=1.5] -- cycle;  
                \fill[memoryColor] (0,1.5) arc[start angle=90, end angle=150, radius=1.5]  -- (0,0) -- cycle;  
                \fill[computeColor] (0,1.5) arc[start angle=90, end angle=140, radius=1.5] -- (0,0) -- cycle;  
            \end{scope}
            
            \begin{scope}[shift={(2,-2)}]
                \node at (-2,0) {\textbf{SNN} (DLIF)};  
                \draw (0,0) circle (1.5);  
                \fill[computeColor] (0,1.5) arc[start angle=90, end angle=95, radius=1.5] -- (0,0) -- cycle;  
                \fill[memoryColor] (0,0) -- (0,1.5) arc[start angle=90, end angle=-265, radius=1.5] -- cycle;  
            \end{scope}
            
            \begin{scope}[shift={(1,-4)}]
                \draw[fill=computeColor, rotate=90] (3.5+2,-2) rectangle (4.9+2,-1.5); 
                \node at (-0.7+2,4.2+2) {\footnotesize Compute};
                \draw[fill=memoryColor,rotate=90] (3.5+2,0+.3) rectangle (4.9+2,0.5+.3);
                \node at (-0.7+1-0.15,4.2+2) {\footnotesize Memory (local)};
                \draw[fill=locmemoryColor,rotate=90] (3.5+2,-1+0.15) rectangle (4.9+2,-.5+0.15);
                \node at (-0.7-0.3,4.2+2) {\footnotesize Memory (distant)};
            \end{scope}
        \end{tikzpicture}    
    \caption{\phantom{x}}
    \label{fig:energy_a}
    \end{subfigure}
    \hspace{1cm}  
    \begin{subfigure}{0.45\textwidth}
        \centering
        \begin{tikzpicture}[scale=0.75]
            \begin{axis}[
                xlabel={$N_{\text{spikes/syn}}$},
                ylabel={$E_{ANN}/E_{SNN}$},
                ymin=0, ymax=10,
                xmin=0, xmax=1,
                samples=200,
                domain=0:120,
                grid=major,
                thick,
                restrict y to domain=0:130,
                legend style={cells={align=left},font=\scriptsize,at={(0.985,0.77)},row sep=10cm}
            ]

            \addplot[black, thick, dotted, domain=0:1] {1};
            \addlegendentry{Transition from\\ $E_\text{ANN}>E_\text{SNN}$ to \\ $E_\text{ANN}<E_\text{SNN}$}
            
            \addplot[red, thick, dashed, domain=0.0005:1] {0.5+1.5/(x)^1.14};
            \addplot[blue, thick, dashed, domain=0.00005:1] {0.06/(x)^1.6};
            \addplot[green, thick, dashed, domain=0.00005:1] {0.3/(x)^1.3};
            
            \node[black, right] at (axis cs:0.35,8.5) {\footnotesize{ANN-base}};
            \draw[->,dashed, black] (axis cs:0.35,8.35) -- (axis cs:0.25,8);
            
            \node[black, right] at (axis cs:0.45,1.5) {\footnotesize{Existing ANN accelerator}};
            \draw[->,dashed, black] (axis cs:0.45,1.5) -- (axis cs:0.35,1.35);
    
            \node[black, right] at (axis cs:0.2,3.5) {\footnotesize{ANN-opt}};
            \draw[->,dashed, black] (axis cs:0.21,3.4) -- (axis cs:0.15,1.4);
   
            \end{axis}
        \end{tikzpicture}        
    \caption{\phantom{x}}
    \label{fig:energy_b}
    \end{subfigure}
    \caption{(a) Relative contributions of (distant/local) memory accesses and compute operations to energy consumption in neurons; ANN-opt is optimized towards local memory usage which is more efficient than distant memory leading to the relative increase of the compute costs. (b) Energy efficiency of DLIF SNNs ($T=1$) relative to ANNs as a function of $N_{\text{spikes/syn}}$ using the AlexNet network topology. ANN-base and ANN-opt denote the worst and best case, respectively, i.e., existing ANN accelerators approach the best case. SNNs achieve superior energy efficiency below $N_{\text{spikes/syn}} \approx 0.4$. Both figures are adapted from \cite{Dampfhoffer22Energy}.}
    \label{fig:energy}
\end{figure}

However, the presented approach does not reflect observations in practice, which are more intricate and dependent on multiple dimensions such as choices regarding hardware, data, implementation, algorithms, etc. For instance, even for networks of the same size, the number of arithmetic operations in SNNs is influenced by the temporal dynamics, meaning that the computation is performed over a certain period in which neurons can fire several (or no) spikes and thereby increase (or decrease) the number of arithmetic operations. Moreover, optimized implementations of ANNs as well as SNNs on dedicated hardware impact the number of arithmetic operations. More importantly, it turns out that arithmetic (compute) operations are not the main source of energy consumption on specialized hardware
; instead, memory accesses and associated communication overhead dominate; see \Cref{fig:energy_a}. However, the number of memory accesses and their associated energy consumption heavily depends on the utilized hardware architecture via dataflow pipelines and the ability to exploit data sparsity. For instance, event-based computation can be leveraged on neuromorphic hardware in ways inaccessible to classical digital hardware \cite{Mehonic2024NeuromorphicRoadmap}. Therefore, an important question is how to assess the energy-efficiency of ANNs and SNNs fairly. 

Certainly, one can \emph{directly measure} the energy consumption of given implementations and compare the results along specific applications and hardware targets. However, this approach has several downsides. Most notably, it is not generalizable as the considered applications are often specific and not representative of real-world AI tasks. Additionally, the approach overlooks potential optimizations of the hardware-software interplay, which would require different strategies for ANNs and SNNs \cite{Lemaire22Energy}.
Therefore, a detailed apple-to-apple comparison is not feasible and 
only the final results can be contrasted. 
Another approach involves \emph{theoretical analysis} of energy consumption to address fundamental questions that are, to some extent, independent of specific implementations by introducing metrics that assess relative energy consumption based on synaptic operations and neural activity. Hence, one can exactly compare ANNs and SNNs in the same pre-determined setting, leaving the problem of defining a fair and meaningful setting for a benchmark. A possible solution is grounding the comparison on low-level and non-specific hardware structures such as CMOS with some reasonable assumptions accounting for memory accesses, which we will not cover in detail. 
Naturally, both approaches are relevant and should be considered as the endpoints of a continuum from `direct' (more specific) to `theory' (more general and abstract) approaches.  

Taking these general thoughts into account, we can refine the previous naive comparison by factoring in memory accesses on digital hardware as suggested in \cite{Dampfhoffer22Energy}. Here, the focus is on dynamic energy consumption, i.e., computation and memory accesses, whereas static energy consumption and communication are disregarded. This is motivated by the main
expected benefit of SNNs---replacement of MACs by ACs, which should be reflected in their dynamic energy consumption.  
Ignoring the application of the activation function $\varphi$ (which for ReLU boils down to a single comparison with zero), the following procedure describes the operations in an artificial neuron according to \eqref{eq:ANNneuron}: $y_i$ is iteratively computed by reading $x_j$, the associated weight $w_{i,j}$ and the current partial sum (psum) from memory followed by a MAC operation and storing the updated psum in memory. 
Denoting the energy of read and write operations of data type $z$ by $\text{ER}_z$ and $\text{EW}_z$, the total energy for computing one pass through an ANN with $N_\text{syn}$ edges is
\begin{equation}\label{eq:ANN_E}
    E_\text{ANN} = N_\text{syn} \times (\text{ER}_\text{input} + \text{ER}_\text{weight} + \text{ER}_\text{psum} + \text{EW}_\text{psum} + E_\text{MAC}). 
\end{equation}
For the $\gS_{\text{DLIF}}$ class, we assume that a spike directly communicates the (memory) addresses of the associated weight and potential, which corresponds to a general event-based (neuromorphic) architecture and not an existing hardware accelerator specific to an SNN model. Hence, according to \eqref{eq:LIF_Energy}, for each incoming spike only the weight and potential must be read and subsequently added via an AC operation. Since a synapse can transmit several spikes, the number of synapses $N_\text{syn}$ is weighted by the average number of spikes per synapse in a network $N_\text{spikes/syn}$ leading to the total energy of passing through a DLIF SNN 
\begin{equation*}
    E_\text{SNN} = N_\text{syn} \times N_\text{spikes/syn} \times (\text{ER}_\text{weight} + \text{ER}_\text{pot} + \text{EW}_\text{pot} + E_\text{AC} ) + N_\text{neur} \times T \times (\text{ER}_\text{pot} + \text{EW}_\text{pot} + E_\text{MAC}),
\end{equation*}
where the second summand accounts for the decaying potential, i.e., the energy for updating the potential of all neurons $N_\text{neur}$ at each of the $T$ time steps corresponding to reading the potential, multiplying it with a constant, and writing back the result. Many fine details are neglected in the analysis as well as contributors ignored such as communication overhead and addressing \cite{Lemaire22Energy}. Moreover, different spiking models vary within their specifications and may be more/less amenable to energy savings. Nevertheless, the derived formula matches the expectation in that decreasing the number of time steps $T$ and increasing the spike sparsity via $N_\text{spikes/syn}$ positively influence the energy consumption of SNNs. This also relates to the previous sections about expressivity, training, and generalization as well as coding by highlighting the need to achieve respective results in the high sparsity and low latency domain to benefit energy efficiency.

While promising, the introduced framework still does not reflect a high-fidelity scenario on digital hardware. ANNs typically are optimized to leverage data reuse by efficient data flow: locally reusing data in several consecutive MAC operations instead of reloading it every time from distant memory, and exploit sparsity in the input via data compression and logic to skip unnecessary MAC operations. A simple example is given by convolutional architectures where weights are reused in multiple synapses, i.e., $N_\text{syn}$ differs from the number of weights, such that memory accesses are minimized and, thus, energy consumption reduced. In contrast, the event-based implementation of SNNs can not leverage data reuse due to the non-flexible and non-predictable order of spike-driven computations; see \Cref{fig:energy_a}. 
Therefore, meaningful comparisons should involve optimized ANN implementations encountered in practice. Incorporating the reuse factor $\text{RF}_z$, which depends on the topology of the ANN architecture/layer and the data $z$ being considered, by weighting the cost of accessing distant memory by the corresponding RF and introducing $\text{ER}^{\text{loc}}$, $\text{EW}^{\text{loc}}$ as the cost of local storage access in \eqref{eq:ANN_E} yields with average input sparsity rate $\gamma \in [0,1]$
\begin{equation*}
    E_\text{ANN} = N_\text{syn} \times (\frac{\text{ER}_\text{input}}{\text{RF}_\text{input}} +  \frac{\text{ER}_\text{weight}}{\text{RF}_\text{weight}} +  \frac{\text{ER}_\text{psum} + \text{EW}_\text{psum}}{\text{RF}_\text{psum}} + \text{ER}^{\text{loc}}_\text{input} + \gamma (\text{ER}^{\text{loc}}_\text{weight} + \text{ER}^{\text{loc}}_\text{psum} + \text{EW}^{\text{loc}}_\text{psum} + E_\text{MAC})),
\end{equation*}
since for a zero input, the MAC, the weight read, and psum read and write in the local memory are saved.

By inserting experimentally derived energy measurements for the read, write, and compute operations on the employed hardware, one can compare the relative energy consumption of ANNs and SNNs for various hyperparameters ($N_\text{syn}$, $N_\text{spikes/syn}$, $N_\text{neur}$, $T$) and identify transitions from $E_\text{SNN} < E_\text{ANN}$ to $E_\text{SNN} > E_\text{ANN}$ (\Cref{fig:energy_b}). 
It is indeed observed in practice that SNNs achieve accuracies on par with ANNs in the $E_\text{SNN} < E_\text{ANN}$, indicating that SNNs can indeed be more energy-efficient than ANNs. However, in the more realistic (optimized implementation) scenario, the energy consumption tilts in favor of ANNs except for low latency and high spike sparsity $N_\text{spikes/syn} \ll 1$ (\Cref{fig:energy}). At the same time, digging deeper into the specific hardware structures can be utilized to optimize the processing and memory management on the SNN side as well at the cost of establishing a fair comparison since ANN and SNN intrinsically benefit from different hardware architectures due to their distinct computation structure \cite{yan2024Energy}. Hence, comparing the energy consumption of ANN and SNNs in frameworks accessible to both computing models, which mostly restricts SNNs to the discretized variations, may favor ANNs without providing a definite answer.

Going beyond digital hardware and accelerators, SNNs also benefit from analog implementations on low-power neuromorphic chips that optimally align the structure of the SNN, not necessarily restricted to the DLIF model, with the underlying hardware. 
Moreover, the previous considerations were based on inference on static data, i.e., data was passed once through the network, however, the natural dynamics of SNNs better align with temporal data produced by neuromorphic sensors, such as event-based cameras, where SNNs show competitive performance to ANNs \cite{ottati2024Accelaration}. Finally, instead of focusing only on inference, one can also include the whole training pipeline in the analysis \cite{yin2022Sata}.

\section*{Acknowledgment}
HB was supported by the Federal Ministry of Education and Research of Germany (BMBF) in the programme of “Souverän.Digital.Vernetzt.”, joint project 6G-life, project identification number 16KISK002.

GK was supported in part by the Munich Center for Machine Learning (BMBF) as well as the German Research Foundation under Grants DFG-SPP-2298, KU 1446/31-1 and KU 1446/32-1.

GK and MS acknowledge support by the Konrad Zuse School of Excellence in Reliable AI (DAAD).

GK, HB, and MS also acknowledge support by the project "Next Generation AI Computing (gAIn)", which is funded by the Bavarian Ministry of Science and the Arts and the Saxon Ministry for Science, Culture and Tourism.

\bibliographystyle{IEEEtran}
\bibliography{IEEEabrv, bibliography}
\end{document}